\documentclass[11pt]{article}
\usepackage{algorithm}
\usepackage{algorithmic}
\usepackage{multirow}

\usepackage[preprint]{acl}

\usepackage{times}
\usepackage{latexsym}

\usepackage[T1]{fontenc}
\usepackage{enumitem}

\usepackage{multirow}
\usepackage[utf8]{inputenc}

\usepackage{microtype}

\usepackage{amsmath}

\usepackage{inconsolata}

\usepackage{graphicx}

\title{\textbf{Sci}entific Knowledge-driven \textbf{D}ecoding \textbf{C}onstraints Improving the Reliability of LLMs}

\author{
 \textbf{Maotian Ma\textsuperscript{1,2*}},
 \textbf{Zheni Zeng\textsuperscript{1*$\dagger$}},
 \textbf{Zhenghao Liu\textsuperscript{3}},
 \textbf{Yukun Yan\textsuperscript{2$\dagger$}},
\\
 \textsuperscript{1}Nanjing University,
 \textsuperscript{2}Tsinghua University,
 \textsuperscript{3}Northeastern University
 \\
 \small{
   \textbf{Correspondence:} \href{mailto:zengzn@nju.edu.cn}{zengzn@nju.edu.cn}, \href{mailto:yanyu.thu@gmail.com}{yanyu.thu@gmail.com}
 }
}

\begin{document}

\maketitle
\begin{abstract}

Large language models (LLMs) have shown strong knowledge reserves and task-solving capabilities, but still face the challenge of severe hallucination, hindering their practical application. Though scientific theories and rules can efficiently direct the behaviors of human manipulators, LLMs still do not utilize these highly-condensed knowledge sufficiently through training or prompting. To address this issue, we propose \textbf{SciDC}, an LLM generation method that integrate subject-specific knowledge with strong constraints. By adopting strong LLMs to automatically convert flexible knowledge into multi-layered, standardized rules, we build an extensible framework to effectively constrain the model generation on domain tasks. Experiments on scientific tasks including industrial formulation design, clinical tumor diagnosis and retrosynthesis planning, consistently demonstrate the effectiveness of our method, achieving a 12\% accuracy improvement on average compared with vanilla generation. We further discuss the potential of LLMs in automatically inductively summarizing highly-condensed knowledge, looking ahead to practical solutions for accelerating the overall scientific research process. All the code of this paper can be obtained\footnote{\url{https://github.com/Maotian-Ma/SciDC}}.
\end{abstract}

\section{Introduction}

While LLMs have demonstrated exceptional processing capabilities across various scenarios, they still cannot guarantee entirely rational decision-making~\cite{huang2025survey}. For instance, in clinical diagnosis, LLMs occasionally exhibit hallucinations, fabricating reasoning basis that contradicts professional experience or medical guidelines, which significantly reduces human trust in them and limits corresponding real-world applications. Ultimately, the data-driven learning pattern and model architecture make LLMs difficult to perform logical reasoning that is completely aligned with theoretical rules. This can lead to a mismatch between the generated content and the physical world.

\begin{figure}[t]
  \includegraphics[width=\columnwidth]{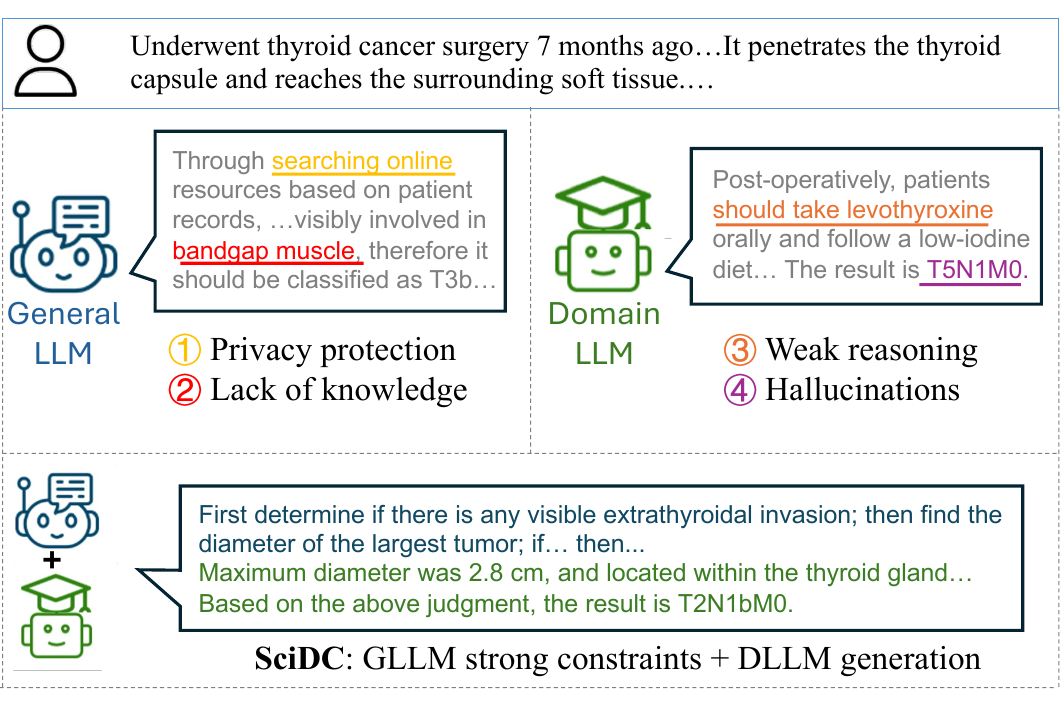}
  \caption{Examples of how general strong LLMs and domain models sometimes fail in specialized tasks.}
  \label{fig:reason}
\end{figure}

To make full use of knowledge like experience rules and professional theories, existing solutions generally fall into two categories: prompt-based methods, which rely on the powerful contextual learning of general models~\cite{li2024enhanced,wang2023knowledge}, similar to the slow thinking process of human students solving problems by following explicit rules; and tuning-based methods, which implicitly parameterize task-specific knowledge~\cite{zeng2025kbalign,zhao2025developing}, similar to the fast thinking process of well-trained experts. Nevertheless, neither of them could not entirely utilize the scientific knowledge to guarantee the logical correctness of LLM reasoning. 

One targeted approach is to conduct post-processing to filter out content that conflicts with scientific knowledge, but this assumes that LLMs can generate useful candidate solutions without prior knowledge guidance. Furthermore, overall efficiency is sacrificed through multiple sampling and verification cycles. To overcome these limitations, we propose a more fundamental solution: integrating knowledge verification directly into the generation process itself. In this paradigm, scientific knowledge acts as a set of decoding constraints, guiding the model to select optimal paths within a feasible region defined by domain expertise. The key challenges thus become: (1) how to systematically formalize diverse knowledge into actionable constraints, and (2) how to efficiently enforce these constraints during the decoding process.


To this end, we introduce a novel \textbf{multi-layered constraint framework} that operationalizes knowledge at different granularities: Top-layer rules govern the macro-structure of reasoning, specifying the necessary steps (e.g., extracting key medical findings before diagnosis). Middle-layer rules enforce complex conditional logic and enable localized regeneration when inconsistencies arise (e.g., adjusting material ratios of formulations based on empirical rules). Bottom-layer rules apply token-level constraints, directly modifying decoder logits to ensure syntactic and numerical feasibility. Implementing this framework requires translating flexible natural language knowledge into standardized, executable rules—a task well-suited for powerful, general-purpose LLMs (GLLMs). Meanwhile, applying these rules during decoding can be efficiently handled by smaller, domain-specific models (DLLMs), which also address data privacy and domain adaptation costs. Thus, we naturally arrive at a \textbf{collaborative architecture}: GLLM parses knowledge into the three-layer constraint set, and DLLM performs the actual generation within these constrained spaces. This approach ensures that domain knowledge is deeply embedded throughout the generation pipeline, enhancing both correctness and practical deployability.


Experiments are conducted on a series of scientific tasks including industrial formulation design, clinical tumor diagnosis and retrosynthesis planning. On different domains and backbone models, our method has shown consistent effectiveness with minimal efficiency loss, improving the accuracy and validity of the generated results. Besides, we further explore the capability of LLM summarizing high-condensed knowledge through hierarchical information organization and reinforcement learning, revealing a possible path to achieve a closed loop in the extraction and use of scientific knowledge.


Our contributions are as follows: (1) We design SciDC, a framework that leverages the knowledge parsing capability of general LLMs while keeping domain data private by using local, smaller models for constrained generation. (2) We conduct a series of experiments across various tasks and backbones, proving the effectiveness of the structured decoding constrains following three layers of standardized rules. An average improvement of 12\% has been achieved compared with vanilla methods. (3) We discuss the closed loop of automated hypothesis generation, verification, and knowledge application, inspiring further LLM for science research.

\section{Related Work}


Due to the in-context learning capabilities of LLMs, it is feasible to achieve knowledge-driven generation by designing prompts or special inference patterns. To integrate knowledge graphs with LLM internal knowledge, models are required to retrieve and construct useful sub-graphs to enhance their reasoning process~\cite{luo2024graph,li2024enhanced,zhao2025knowpath}. In scientific applications, notable examples from the healthcare domain~\cite{jiang2024reasoning} adopt similar methods to reduce hallucinations. Adapting LLMs to specific domains through parameter tuning is also a common approach. Some work emphasizes the model autonomy and transfer learning efficiency~\cite{zeng2025kbalign}, and some tries to combine the retrieval-augmented generation (RAG)~\cite{lewis2020retrieval} setting with targeted training for domain knowledge utilization~\cite{zhang2024raft}. Domain LLMs, such as biomedicine and chemistry models, are also emerging in large numbers, possessing even stronger reserves of professional knowledge~\cite{goyal2024healai,zhao2025developing}. However, these methods do not fully guarantee the accuracy of the content generated by the models.


More stringent constraints based on knowledge are typically applied to tasks that generate structured content such as code. Early work selects few-shot examples from traditional training dataset and shows the effectiveness of decoding constraints~\cite{poesia2022synchromesh}. More tasks such as playing chess~\cite{ma2025logically} are also explored, demonstrating the validity of hand-crafted rules. Recently, researchers focus more on combining the hard constraints with the original knowledgeable distribution of LLMs, using better aligned method to avoid interfering with the normal LLM reasoning process~\cite{park2024grammar,banerjee2025crane}. One representative scenario is in-context knowledge editing, in which previous works try to adopt the contrast logits to enhance the newly added knowledge in reasoning~\cite{bi2024decoding}, or utilize step-wise knowledge selection to improve the reasoning process instead of logits-level mandatory constrains~\cite{wang2024deepedit}. When partial backtracking is allowed, some works conduct iterative structured generation~\cite{ugare2024itergen}, or even train the model to detect hallucinations and re-sample in an online manner~\cite{wu2025generate}.

It is a widely-adopted strategy to conduct collaboration between large and relatively small models. As claimed in the surveys~\cite{chen2024role,wu2025knowledge,chen2025survey}, common routines include knowledge distillation, tool calling and so on. Privacy protection and efficiency improvement are key motivations for conducting collaboration. In this work, we would like to further emphasize that, due to the lightweight, flexible, and domain-knowledge-adaptable nature of downstream models, collaborative approaches might be inherently more effective than working alone.

\section{Methodology}

\begin{figure*}[ht]
\centering
\includegraphics[width=0.84\linewidth]{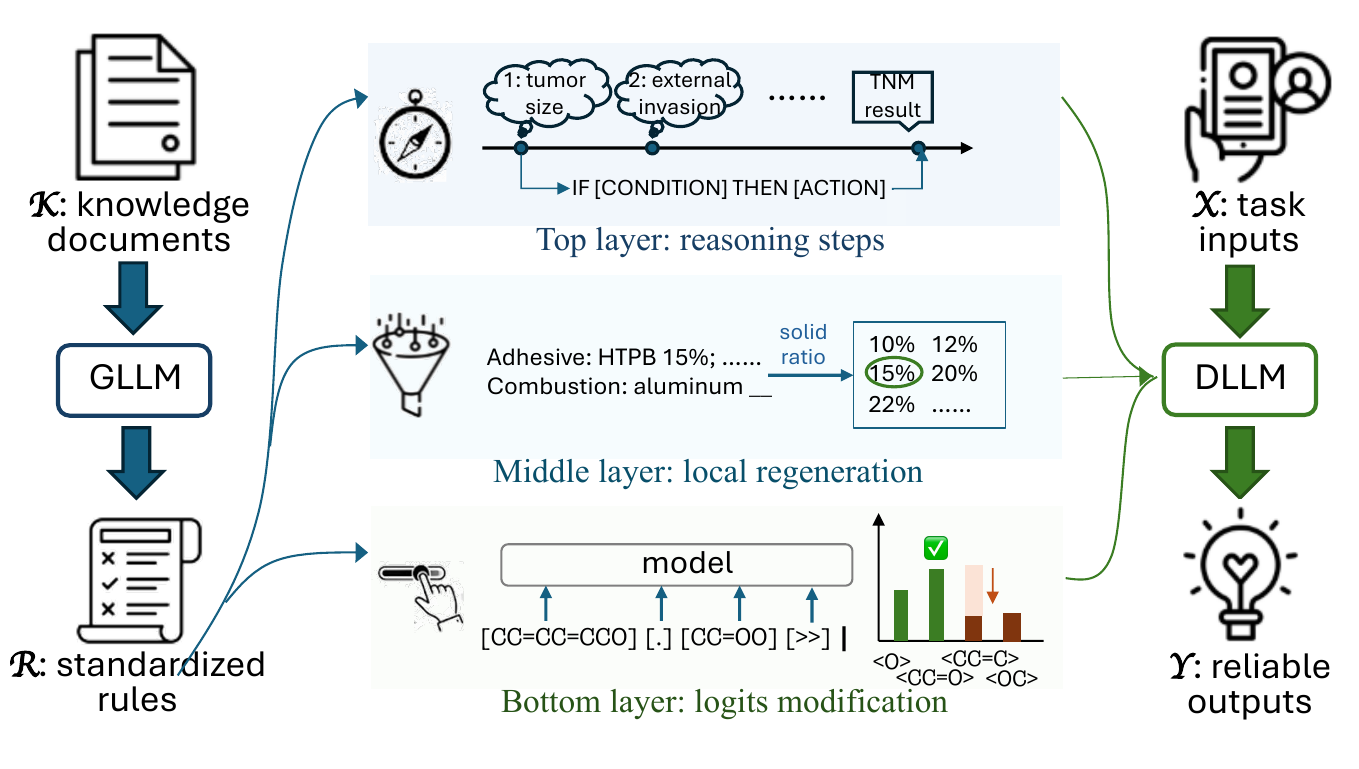}
\caption{Overall framework of SciDC. General LLM transforms knowledge documents into standardized rules, which are divided into 3 layers constraining the reliable generation of domain LLM.}
\label{fig:framework}
\end{figure*}

\subsection{Task Setup}
We abstract the LLM generation task in the scientific domain into this process:
$(\mathbf{X}, \mathbf{K}) \rightarrow \mathbf{Y}$, where $\mathbf{X}$ is the input context providing initial conditions (e.g., patient records, target product molecules), and $\mathbf{Y}$ represents the desired output sequence like clinical diagnosis and synthesis pathway. $\mathbf{K}$ is professional domain knowledge such as scientific theories or empirical experience, usually described in flexible natural language. Considering that LLMs suffer from severe hallucination especially on subject-specific scenarios, our basic target is to ensure that $\mathbf{Y}$ strictly adheres to $\mathbf{K}$.

LLM generation is essentially a decoding search within a high-dimensional semantic space. Domain knowledge helps eliminate some erroneous results, thereby constraining the feasible region of decoding.  We define $\mathcal{S}$ as the theoretical, unrestricted space of all possible $\mathbf{Y}$. $\mathbf{K}$ imposes a set of rules and principles that act as an eligibility filter, which partitions $\mathcal{S}$ to yield the feasible sub-space:
\[
\mathcal{S}_{\mathbf{K}} = \{\mathbf{Y} \in \mathcal{S} \mid \mathbf{Y} \text{ strictly complies with } \mathbf{K}\}
\]
Therefore, the domain models' objective is to select the optimal sequence from the constrained region to ensure scientific soundness:
\[
\mathbf{Y}^* = \arg \max_{\mathbf{Y} \in \mathcal{S}_{\mathbf{K}}} P(\mathbf{Y} \mid \mathbf{X}, \mathbf{K})
\]

For ease of interaction, $\mathbf{K}$ is typically provided with natural language, while its flexibility poses a challenge to accurately defining $\mathcal{S}_{\mathbf{K}}$. Naturally, a transformation mechanism is needed to extract a set of standardized rules $\mathbf{R}$ from $\mathbf{K}$. 

The transformation stage requires a strong generalization capability, and the subsequent generation stage requires a solid foundation in subject knowledge and data privacy protection. Therefore, we first employ a strong general model (GLLM) as a knowledge compiler $\mathbf{K} \rightarrow \mathbf{R}$, and then adopt a locally-deployable domain model (DLLM) for the specific downstream task $(\mathbf{X}, \mathbf{R}) \rightarrow \mathbf{Y}^*$. The collaboration framework is shown in Fig.~\ref{fig:framework}.

\subsection{Standardized Rule Transformation}



The GLLM organizes the extracted knowledge into three distinct layers, creating a multi-grained interface for the semantic space constraints:

\begin{itemize}[leftmargin=*,labelwidth=0pt,labelsep=4pt]
\item \textbf{Bottom-Layer Rules ($R_{\text{B}}$):} Token-level constraints that map directly into local modifications of the decoder logits. 

\item \textbf{Middle-Layer Rules ($R_{\text{M}}$):} Logic-level constraints involving complex multi-hop dependencies that cannot be reduced to simple token probabilities. Conditional checks and local regeneration are conducted to ensure logical consistency.

\item \textbf{Top-Layer Rules ($R_{\text{T}}$):} Structure-level constraints that define the overall reasoning framework into specific sub-task steps. They govern the macro-organization of the output, ensuring the model to follow a specific paragraph sequence.
\end{itemize}
Here we take the rule transformation of formulation design in Algorithm~\ref{alg:formulation} as an instance. Expert knowledge about industrial formulation is mapped onto three granularities of control. $R_\textbf{T}$ establishes the global reasoning workflow (e.g., calculate plasticizer ratio and then choose curing agent). $R_\textbf{M}$ implements conditional logic through multi-hop dependency checks (e.g., matching the sum of curing agent to a ratio range). Finally, $R_\textbf{B}$ utilizes local enumerate parameters (e.g., select an option from given pools) to ensure the final output complies with expert experience.


\begin{algorithm}[h]
\caption{Constrains instance on formulation design task}
\label{alg:formulation}
\begin{algorithmic}[1]
\STATE \COMMENT{\textbf{$R_T$: Step sequence enforcement}}
\STATE \textit{\# 8-step structured reasoning as per $R_T$}
\STATE lm += "Step 1: Extract plasticizer ratio..."
\STATE lm += gen(name="step\_1", max\_tokens=256)
\STATE \textcolor{gray}{$\vdots$}
\STATE lm += "Step 8: Output optimized formula..."
\STATE lm += gen(name="adjusted\_formula", max\_tokens=2048)

\STATE \COMMENT{\textbf{$R_M$: Conditional logic enforcement}}
\STATE \textit{\# Validation loop with constraint checking}
\STATE \textbf{while} not constraint\_satisfied:
\STATE \quad lm += gen(name=f"adjusted\_formula", ...)
\STATE \quad \textcolor{gray}{[... multi-hop validation...]}
\STATE \textit{\# Dynamic options of calculated values}
\STATE current\_ratio = float(lm["current\_ratio"])
\IF{current\_ratio $\geq$ 2.5}
    \STATE options = ["reach the upper limit"]
\STATE \textcolor{gray}{$\vdots$}
\ELSE
    \STATE options = ["not yet", "close to the limit"]
\ENDIF
\STATE \COMMENT{\textbf{$R_B$: Token-level constraints}}
\STATE \textit{\# Option selection constraint}
\STATE lm += select(options=options)
\STATE \textcolor{gray}{$\vdots$}
\STATE \textit{\# Direct logit masking for numerical ranges}
\STATE lm += gen(name="binder", regex=r"\d+\.?\d*", max\_tokens=10)
\STATE \textcolor{gray}{$\vdots$}

\STATE Apply mask to logits: $l'[j] = l[j]$ if $j \in$ valid\_tokens else $-\infty$
\end{algorithmic}
\end{algorithm}


\subsection{Decoding-Constrained Generation}

The constrained generation process is implemented by integrating $\mathbf{R}$ at different stages of DLLM decoding loop, leading to the multi-layered constraint framework. Through the extracted rule file and hand-crafted templates, the constrained generation code can be automatically designed.

The \textbf{bottom-layer constraint} is the strongest form of control, operating at the fundamental logit vector $\mathbf{l}$ of the DLLM to ensure that the immediate token generated adheres to $R_\textbf{B}$.


For the decoding step $i$, DLLM is allowed to choose $y_i$ from a range of valid tokens $\mathcal{V}_i$, which is determined by the syntax, enumeration, and numerical range requirements from $R_\textbf{B}$. This is achieved by applying a mask to logit $\mathbf{l}_i$:
$$\mathbf{l}'_i[j] =
\begin{cases}
\mathbf{l}_i[j], & \text{if token } j \in \mathcal{V}_i \\
-\infty, & \text{if token } j \notin \mathcal{V}_i
\end{cases}$$

This operation is an efficient, matrix-level process, and has negligible impact on generation efficiency. Resulting probabilities of those invalid tokens are set to zero after softmax. Thus, the selected token $y_i^*$ may differ from the model's original preference $y_i$, but it is guaranteed to satisfy the scientifically feasible condition.




The \textbf{middle-layer constraint} handles more complex conditional rules that cannot be simply reduced to logit masking, conducting local check to cross-token segments to see whether they satisfy $R_{\text{M}}$. Once an unreasonable situation is detected (e.g., ${w}_{neg}=\{y_{i-2}, y_{i-1}, y_i\}$ is a component that conflicts with the previously determined ones), DLLM is asked to regenerate from $y_{i-2}$ in a loop, until the sample result is correct (e.g., ${w}_{pos}=\{y^*_{i-2}, y^*_{i-1}\}$ is allowed). Notice that the backtracking length in SciDC is not fixed but dynamically determined by the semantic units defined in the GLLM rules. Inevitably, the process of sampling and judging multiple times will increase the time cost to some extent. 






The \textbf{top-layer constraint} targets the macro-structure of $\mathbf{Y}$ and operates sentence-level or even paragraph-level generation. By enforcing the sequential logic in $R_{T}$, it ensures that the Chain-of-Thought (CoT) of DLLM remains consistent with established scientific protocols or guidelines. Here we provide a specific example: $p_0=\{y_0, ..., y_k\}$, $p_1=\{y_{k+1}, ..., y_m\}$, $p_2=\{y_{m+1}, ..., y_n\}$ and $p_3=\{y_{n+1}, ..., y_q\}$ consist of the whole response $\mathbf{Y}$, in which $p_0$ and $p_2$ are the pre-defined formatting fields or prompting issues (e.g., "we need to first figure out the maximum tumor diameter"), thus DLLM is required to complement $p_1$ and $p_3$ with these implicit hints. Since we mandates the overall output into a relatively structured format, this can improve both professional consistency and seamless integration with downstream systems.


We provide a concrete implementation of the multi-layered constraint framework in Algorithm~\ref{alg:formulation}, where knowledge-aware rules generated by the GLLM are described by executable code. \( R_T \) enforces a predefined step-by-step reasoning sequence by prompt design (lines 2–8), ensuring the overall narrative follows established scientific protocols. \( R_M \) are realized via conditional validation loops and dynamic option selection (lines 10–20), enabling multi-hop dependency checks and local regeneration when logical inconsistencies arise. Finally, \( R_B \) are applied through direct logit masking and regex-based generation (lines 22–26), guaranteeing that numerical values and categorical choices remain within scientifically feasible ranges. Through such automatic code generation, domain knowledge is systematically embedded into each stage of decoding, allowing the model to produce outputs that are both linguistically coherent and experientially grounded.

\section{Experiment}

To evaluate the effectiveness of our method, we conduct extensive experiments across three distinct domains: industrial formulation design, clinical tumor diagnosis, and chemical retrosynthesis planning. These tasks represent high-stakes decision-making scenarios that require both specialized domain knowledge and logical reasoning.

\subsection{Datasets and Evaluation Metrics}

We evaluate our framework across three representative domains, and the adopted datasets cover open-source / private scenarios,  real / simulated data, and relatively simple / complex tasks. 

(1) \textbf{Formulation design}: The model optimizes a formulation of 10-20 components (substance, amount and other details) to meet specific requirements (a higher plasticizing ratio). The knowledge document, totaling 1.7k tokens, provides empirical guidelines on the combination and proportion of substances in such kind of formulations, compiled by professionals. 458 formulation samples are provided, and models are evaluated on validity (adherence to guidelines) and success rate (achievement of specified requirements).

(2) \textbf{Tumor diagnosis}: The model reads 200 virtual medical records (verified by clinical professionals) of patients with thyroid cancer and answers their TNM stage of malignancy. The 500-token TNM staging guideline for thyroid cancer is primarily based on the 8th edition of the American Joint Committee on Cancer criteria for thyroid cancer~\cite{lamartina2018ajcc}. Validity (a complete and clinically reasonable assessment) and exact match (the result is entirely correct) are reported.

(3) \textbf{Retrosynthesis}: The model recommends feasible reactions for given products. To facilitate automated evaluation, we limit it to a single-step reaction, and utilize 25,214 official reaction templates from the USPTO-460k dataset, as provided in the standard configuration of AiZynthFinder~\cite{thakkar2024aizynthfinder}. These templates, along with professional functions for template search and matching, constitute the knowledge document for this task. We randomly select 201 reactions from USPTO-full of which the products are unseen in the template source, and evaluate the hit@1 and reaction validity of the recommended results. 



\subsection{Models and Baselines}


For the upstream GLLM, we observe several popular strong models and adopt Claude-3.5-Sonnet~\cite{anthropic2024claude35sonnet} for its superior performance in code generation and structured output adherence~\cite{kaspero2024evaluating, anthropic2024claude}. We also include it, along with other representative models (e.g., GPT-5-chat~\cite{openai2025gpt5}), as baselines for comparison. Details including prompt templates are provided in Section~\ref{sec:prompt}.


For the downtream DLLM, we uniformly try Qwen3-14B and Qwen3-4B~\cite{yang2025qwen3}, which are widely tested open-source models with multi-domain capabilities. We also deploy ChemDFM-v1.5-8B~\cite{zhao2025chemdfm} as a representative domain-specific LLM for retrosynthesis planning, validating the generalization of SciDC.


In the following experiments, our base setting is to provide a concise task prompt and knowledge document $\mathbf{K}$, and allow the models to conduct a long deep thinking before generating final response. The baseline \textit{w/o $K$} removes the knowledge document from model inputs, and \textit{w SciDC} applies our framework. Given the large operational space, lack of typical samples, and the fact that relevant knowledge has already been provided, we do not provide few-shot retrieval samples in the context.


\subsection{Expert-in-the-loop Rule Verification}
To ensure the reliability and stability of the automatically synthesized constraints, we conduct an interactive human verification for the generated rule codes. That is, a conversation between GLLM and a human expert, in which the model describes what the code is trying to do in purely natural language (e.g., "...the program uses the tumor size, location, and extent of invasion to provide possible T candidate values: smaller than 1cm and within the thyroid gland → T1a..."), and the human provides suggestions within 2 turns. In this way, the human expert does not need to understand code language, but can help verify the system. 

Table~\ref{gllm_eval} demonstrates the blind human evaluation results (0--5 scale) of the initial generated codes in tumor diagnosis. The overall performance of Claude is slightly better than GPT-5, which is the reason why we choose it as GLLM. Meanwhile, errors also exist in the initial generated codes. For compilation correctness, few results apply wrong regex arguments, which can be corrected in human interaction. For logical integrity, the problem occasionally arises when extracting key information. In such cases, the appropriate options should be provided (no distant transfer vs. distant transfer exists) rather than directly requiring a determination of the period (M0 vs. M1). For efficiency, the problem with all the results is that redundant judgments are made for clear jump relationships. For example, when "no distant transfer" is selected, the program correctly restricts it to only one option M0, but still requires the model to generate an analysis of the reasons, which leads to a decrease in efficiency. Nevertheless, in our setting, with the expert-in-the-loop verification, GLLM can elevate the constraint codes to a nearly perfect quality.

\begin{table}[ht]
\centering
\caption{Human evaluation for GLLM-generated constraints. We report the average scores of 5 samples. }
\label{gllm_eval}
\resizebox{0.9\linewidth}{!}{
\begin{tabular}{lccc}
\hline
\textbf{Model} & \textbf{Correctness} & \textbf{Integrity} & \textbf{Efficiency} \\
\hline
GPT-5 & 4.6 & 4.0 & 3.0 \\
Claude & 4.6 & 4.6 & 2.6 \\
\textbf{SciDC} & \textbf{5.0} & \textbf{4.8} & \textbf{3.2} \\
\hline
\end{tabular}}
\end{table}

\subsection{Result Analysis}



\begin{table*}[ht]
\centering
\resizebox{0.9\linewidth}{!}{
\begin{tabular}{l|cc|cc|cc|c}
\hline
\multirow{2}{*}{\textbf{Method}} & \multicolumn{2}{|c|}{\textbf{Formulation design}} & \multicolumn{2}{c|}{\textbf{Tumor diagnosis}} & \multicolumn{2}{c|}{\textbf{Retrosynthesis}}  & \multirow{2}{*}{\textbf{Overall}}\\
 & validity & success rate & validity & exact match & validity & hit@1 &  \\
\hline
GPT-5 & \underline{92.3} & \underline{92.2} & 98.5 & \underline{94.5} & 64.7 & 31.8 & \underline{72.8} \\
Claude-3.5 & 88.6 & 82.6 & \underline{100.0} & 88.5 & \underline{92.8} & \underline{47.3} & 72.8 \\
\hline
Qwen3-4B & 43.4 & 43.0 & 97.3 & 59.2 & 84.6 & 25.4 & 42.5\\
\ \ \textit{w/o $K$} & 56.8 & 56.8 & 94.3 & 4.0 & 63.5 & 0 & 20.3 \\
\ \ \textit{w SciDC} & 71.0 & 43.4 & 100.0 & 66.7 & 100.0 & \textbf{52.2} & 54.1\\
\ \ \ \ {\color{blue} $\Delta$} &{\color{blue}+27.6} &{\color{blue}+0.4} & {\color{blue}+2.7} & {\color{blue}+7.5} & {\color{blue}+15.4}& {\color{blue}+26.8} & {\color{blue}+11.6}\\
\hline
Qwen3-14B & 50.9 & 50.4 & 79.7 & 72.0 & 78.1 & 31.8 & 51.4 \\
\ \ \textit{w/o $K$} & 18.4 & 18.4 & 74.3 & 36.5 & 48.5 & 0 & 18.3 \\
\ \ \textit{w SciDC} & \textbf{75.5} & \textbf{68.3} & \textbf{100.0} & \textbf{79.5} & \textbf{100.0} & 41.3 & \textbf{63.0} \\
\ \ \ \ {\color{blue} $\Delta$} &{\color{blue}+24.6} & {\color{blue}+17.9}& {\color{blue}+20.3} & {\color{blue}+7.5} & {\color{blue}+21.9} &{\color{blue}+9.5} & {\color{blue}+11.6}\\
\hline
\end{tabular}}
\caption{Main results for SciDC framework on 3 distinct domain. We report average scores for 3 random seeds. \textbf{Overall} shows the average accuracy of the three tasks. $\Delta$ stands for delta values after applying our framework. }
\label{tab:main}
\end{table*}

Main results for our experiments are shown in Table~\ref{tab:main}. Generally, SciDC can bring an obvious and consistent improvement regardless of specific domains or backbones. Simply by providing the knowledge documents and task requirements to our framework, models can consistently achieve performance gains. The difference between with and without $\mathbf{K}$ demonstrates the importance of knowledge introduction, while these knowledge documents are still not being fully utilized. By using Claude-3.5-Sonnet as a one-time knowledge compiler, a locally-deployed model can achieve comparable performance under our framework.

\textbf{Task difficulty}. It is worth mentioning that the extent to which the SciDC framework can be helpful also depends on the difficulty of the task. In tumor diagnosis task, the auto-generated virtual medical records are shorter and clearer than in real cases, and the backbone model itself could reach a relatively high score. However, when tested on a small number of real medical records, Qwen3-14B can only get 33.5\% exact match. With the assistance of SciDC, the score raises to 45.1\% (+11.6). The intervention of professionals (such as adjusting the prompts appropriately based on performance) can even ultimately help the score reach over 70\%, approaching the effect of GPT-5-chat on this data. This suggests that SciDC has greater potential to be effective in more challenging professional contexts.

Results on the retrosynthesis task further support this observation. If professional tool calling not allowed, neither general LLMs nor domain models can directly recommend reasonable reactions for the given products. However, our decoding constraint framework still applies in this case by converting the reaction template knowledge (rather than direct search function calls) into structured decoding constraints. This improves the hit score of ChemDFM from 16.2\% to 29.0\%.

\textbf{Capability of domain-specific models}. Related results are shown in Table~\ref{tab:chemical}. Results for pure template-based method can demonstrate that introducing capabilities of large models is necessary for retrosynthesis planning. Owing to its good expertise in corresponding tasks, ChemDFM-8B shows a satisfying validity and accuracy on retrosynthesis, and SciDC further helps refine the results. This echoes our earlier point that tuning with a large amount of domain-specific corpus cannot make the model completely conform to the rules of the subject, but our method and the knowledge possessed by the domain model can demonstrate a synergistic effect. Meanwhile, we actually test some other domain models, and find that an unavoidable challenge is the balance between general capabilities (e.g., in-context learning and instruction following) and domain knowledge memorizing. Though these models perform well in tasks such as knowledgeable question-answering, they cannot easily surpass Qwen3-14B on our tasks since the formats are not common in domain adaptation learning. Once the loss of general capabilities exceeds a certain limit, providing knowledge materials, or restricting the reasoning process, may even degrade their performances in some cases.


\begin{table}[ht]
\centering
\resizebox{0.75\linewidth}{!}{
\begin{tabular}{lcc}
\hline
\textbf{Method}      & \textbf{validity} & \textbf{hit@1}   \\
\hline
\textbf{Template-based} &100 &31.8  \\
\textbf{ChemDFM-8B}       &  88.5 & 6.0  \\
\ \ \textit{w/o $K$} & 51.5& 0 \\
\ \ \textit{w SciDC} &100 &54.4  \\
\ \ \ \ {\color{blue} $\Delta$} & {\color{blue}+11.5} & {\color{blue}+22.6}  \\
\hline
\end{tabular}}
\caption{Retrosynthesis planning performance of our framework on domain specific model.}
\label{tab:chemical}
\end{table}

\begin{table}[ht]
\centering
\resizebox{0.85\linewidth}{!}{
\begin{tabular}{lccc}
\hline
\textbf{Method}      & \textbf{validity} & \textbf{accuracy}  & \textbf{Step} \\
\hline
\textbf{Ours}       &  75.5 & 68.3 &18 \\
\ \ \textit{w/o $R_T$} &63.8 &24.0 &13 \\
\ \ \textit{w/o $R_M$} & 76.0& 38.6& 15\\
\ \ \textit{w/o $R_B$} & 36.9& 25.3& 8\\
\ \ \textit{w/o all} & 50.9 & 50.4 & -\\ 
\hline
\end{tabular}}
\caption{Ablation results for Qwen3-14B on the formulation design task. \textbf{Step} refers to number of executed constrained steps.}
\label{tab:ablation}
\end{table}

\textbf{Inference efficiency}. Consider that the pre-defined reasoning framework and the local re-generation setting may impair overall efficiency, we analyze the results of the standard generation and our setting. In formulation design, vanilla reference requires 3.6k tokens per piece, while ours gets 4.2k tokens, with 0.8 times of regeneration on average. In retrosynthesis, the corresponding token lengths are 1.9k vs. 2.3k (with 2.5 times of regeneration). As can be seen, the efficiency loss in terms of generation length remains within a relatively acceptable range. However, it should be noted that in the current code implementation, repeated calls to the generation function lead to some redundant data loading and calculations, thus our actual time cost is obviously higher than baselines ($\sim$ 3 times in practice). This can be considered as an optimization direction in actual deployment.

\textbf{Ablation study}. We conduct ablation study of Qwen3-14B on the formulation design task. As shown in Table~\ref{tab:ablation}, removing each layer of rule constraints can cause a decrease of scores, among which the top and bottom layer plays more important role, both in terms of metric improvement and number of code steps. Though current LLMs have strong instruction following capabilities, they still do not exhibit logical consistency of required format. Meanwhile, the vanilla reasoning pipeline of the model also has considerable room for improvement. As for the middle layer constraints, they provide a considerable assistance, but at the cost of reduced overall efficiency.

\subsection{Case and Generalizability Studies}

\begin{figure*}[ht]
\centering
\includegraphics[width=\linewidth]{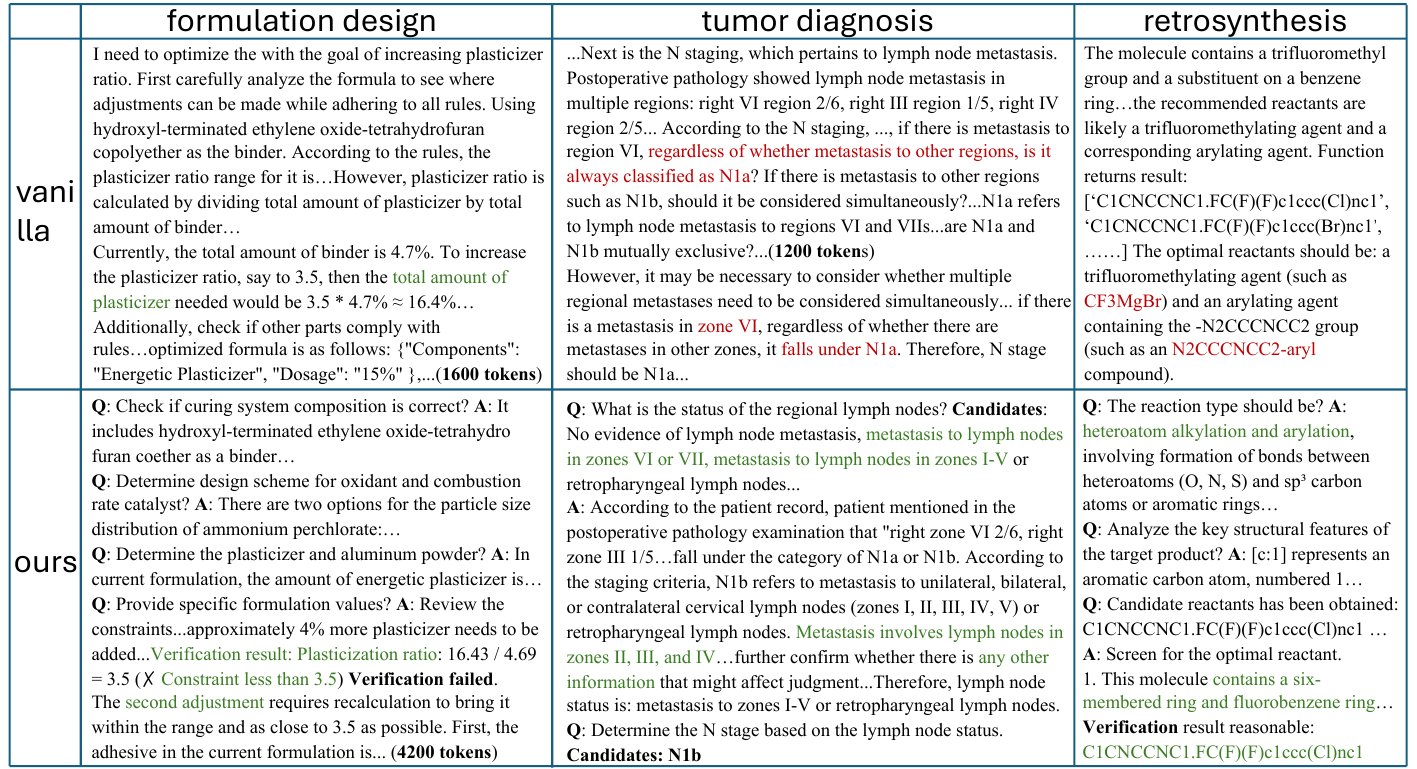}
\caption{Case study comparing the vanilla prompt-based method and SciDC results.}
\label{fig:case}
\end{figure*}


Figure~\ref{fig:case} illustrates typical cases contrasting vanilla generation with our constrained framework. In tumor diagnosis, Qwen3-14B makes a rash decision on lymph node metastasis (N category), hesitating for an extended period but eventually incorrectly suggesting N1a, despite the clinical urgency of N1b per guidelines. SciDC prevents this by enforcing a top-layer reasoning sequence (assess nodes, then stage) supplemented by middle-layer logic checks (e.g., involvement of zones I-V mandates $\geq$ N1b), ensuring alignment with expert protocols.


In retrosynthesis planning where precision is critical, bottom-layer constraints become more prominent. Vanilla generation erroneously proposes invalid reactant candidates, and ultimately generate implausible SMILES strings. Our framework addresses this by first applying a classify step derived from knowledge rules, which then activates specific token-level constraints during SMILES generation. Consequently, the proposed reaction pathway and the final molecular representation remain chemically plausible and syntactically correct.

Meanwhile, we have already discussed the efficiency trade-off in our approach, while some extreme cases appear in complex formulation design tasks. We present a case where both vanilla generation and our framework successfully adjust a plasticizer ratio. However, while the vanilla method uses ~1.6k tokens, our framework produces a >4.2k token trace, meticulously following the sequence mandated by \(R_T\) and iteratively validating intermediate parameters against \(R_M\). This results in a noticeable decrease in inference efficiency that is intrinsic to the method's design: the extended, structured reasoning process guarantees that every step of the output adheres to expert-derived constraints, thereby maximizing professional consistency and auditability at the expense of raw speed. This demonstrates that our framework prioritizes reliable, knowledge-grounded generation over unconditional generative fluency.

It is important to note that the general transferability across different tasks is precisely the strength of our method. Although SciDC is initially designed and evaluated in scientific reasoning, it also demonstrates structured rule-based reasoning capabilities for a wider range of natural language processing tasks. To verify this, we adopt LegalBench~\cite{guha2023legalbench}, a benchmark for evaluating the structured legal knowledge reasoning capabilities. One task with significant real-world application is hearsay, i.e., determining whether a given description of evidence is hearsay evidence. Section~\ref{legalbench} presents the results of this experiment, where our framework, through hierarchical rule generation of the legal definition of hearsay evidence, successfully outperforms GLLMs by the small backbone Qwen3-14B.

\section{Discussion}

While SciDC has demonstrated superiority over multiple tasks, we must still acknowledge the existing problems. For instance, while algorithmic and rule-based knowledge (e.g., TNM staging) is easily compiled, tacit domain knowledge poses a significant challenge. In a failure case within the formulation design task, the GLLM increased the plasticizer ratio from 3.1\% to 4.76\% to satisfy a specific concentration rule but failed to adjust other components accordingly. This resulted in a total mass fraction of over 100\%, violating the physical law of mass conservation. This demonstrates that for SciDC to be effective, all fundamental physical and heuristic constraints must be explicitly codified, as models often fail to internalize unwritten domain knowledge.

Further, we turn to a preliminary discussion about how explicit written knowledge is acquired. With the development of LLMs, we wonder whether they could join the process of proposing new scientific hypotheses, to automate the entire process from knowledge generation to application along with SciDC, inspiring further scientific development. We have conducted a rough verification (shown in Section~\ref{sec:appendix}) of LLM raising scientific hypotheses, while do not reach satisfying results. Future explorations may include the automatic knowledge discovery, to realize the human-in-the-loop end-to-end intelligent research system.

\section{Summary}

In this work, we propose SciDC, an automatic framework converting flexible scientific knowledge into multi-layered, formalized decoding constraints for LLM generation. A series of experiments have demonstrated the effectiveness and portability of our framework. Without any parameter tuning or hand-crafted modification, SciDC could collaborate stronger LLMs with locally-deployed domain models and achieve better performances. This makes it particularly suitable for privacy-sensitive downstream applications. In the future, we will explore the automatic knowledge proposing to form intelligent loop for scientific applications.

\section*{Limitations}

Our framework, while effective, has several limitations. Conceptually, the core idea of orchestrating larger and smaller LLMs is not novel; our primary contribution lies in the formalization of this process into a systematic, reusable pipeline for knowledge-constrained generation. On the technical side, our current implementation prioritizes generality over optimal efficiency. For instance, the integration with inference engines like vLLM involves multiple separate calls, and we do not preserve intermediate hidden states across decoding steps, which introduces unnecessary computational overhead. Significant optimizations at the decoder architecture level remain possible. Furthermore, the framework's evaluation is constrained by the current scarcity of high-quality, domain-specialized small models. Consequently, its performance on certain tasks does not surpass that of generalist LLMs like Claude, highlighting the dependency of our approach on the capabilities of the downstream DLLM. Future work will focus on architectural refinements for efficiency and broader validation across domains as more capable specialized models become available.

\section*{Acknowledgments}

This study was supported by data and expertise provided by: Eng. Siyu Ma from Shanghai Space Propulsion Technology Research Institute; Dr. Xin Zhang from Department of Nuclear Medicine, Peking Union Medical College Hospital.


\bibliography{custom}

@article{guha2023legalbench,
  title={Legalbench: A collaboratively built benchmark for measuring legal reasoning in large language models},
  author={Guha, Neel and Nyarko, Julian and Ho, Daniel and R{\'e}, Christopher and Chilton, Adam and Chohlas-Wood, Alex and Peters, Austin and Waldon, Brandon and Rockmore, Daniel and Zambrano, Diego and others},
  journal={Advances in neural information processing systems},
  volume={36},
  pages={44123--44279},
  year={2023}
}

@article{luo2024graph,
  title={Graph-constrained reasoning: Faithful reasoning on knowledge graphs with large language models},
  author={Luo, Linhao and Zhao, Zicheng and Haffari, Gholamreza and Li, Yuan-Fang and Gong, Chen and Pan, Shirui},
  journal={arXiv preprint arXiv:2410.13080},
  year={2024}
}

@article{zhao2025knowpath,
  title={KnowPath: Knowledge-enhanced Reasoning via LLM-generated Inference Paths over Knowledge Graphs},
  author={Zhao, Qi and Yang, Hongyu and Song, Qi and Yao, Xinwei and Li, Xiangyang},
  journal={arXiv preprint arXiv:2502.12029},
  year={2025}
}

@inproceedings{zeng2025kbalign,
  title={KBAlign: Efficient Self Adaptation on Specific Textual Knowledge Bases},
  author={Zeng, Zheni and Chen, Yuxuan and Yu, Shi and Wang, Ruobing and Yan, Yukun and Liu, Zhenghao and Wang, Shuo and Han, Xu and Liu, Zhiyuan and Sun, Maosong},
  booktitle={Findings of the Association for Computational Linguistics: EMNLP 2025},
  pages={13519--13532},
  year={2025}
}

@article{zhang2024raft,
  title={Raft: Adapting language model to domain specific rag},
  author={Zhang, Tianjun and Patil, Shishir G and Jain, Naman and Shen, Sheng and Zaharia, Matei and Stoica, Ion and Gonzalez, Joseph E},
  journal={arXiv preprint arXiv:2403.10131},
  year={2024}
}

@inproceedings{goyal2024healai,
  title={Healai: A healthcare llm for effective medical documentation},
  author={Goyal, Sagar and Rastogi, Eti and Rajagopal, Sree Prasanna and Yuan, Dong and Zhao, Fen and Chintagunta, Jai and Naik, Gautam and Ward, Jeff},
  booktitle={Proceedings of the 17th ACM International Conference on Web Search and Data Mining},
  pages={1167--1168},
  year={2024}
}

@article{zhao2025developing,
  title={Developing ChemDFM as a large language foundation model for chemistry},
  author={Zhao, Zihan and Ma, Da and Chen, Lu and Sun, Liangtai and Li, Zihao and Xia, Yi and Chen, Bo and Xu, Hongshen and Zhu, Zichen and Zhu, Su and others},
  journal={Cell Reports Physical Science},
  volume={6},
  number={4},
  year={2025},
  publisher={Elsevier}
}

@article{jiang2024reasoning,
  title={Reasoning-enhanced healthcare predictions with knowledge graph community retrieval},
  author={Jiang, Pengcheng and Xiao, Cao and Jiang, Minhao and Bhatia, Parminder and Kass-Hout, Taha and Sun, Jimeng and Han, Jiawei},
  journal={arXiv preprint arXiv:2410.04585},
  year={2024}
}

@inproceedings{li2024enhanced,
  title={An enhanced prompt-based llm reasoning scheme via knowledge graph-integrated collaboration},
  author={Li, Yihao and Zhang, Ru and Liu, Jianyi},
  booktitle={International Conference on Artificial Neural Networks},
  pages={251--265},
  year={2024},
  organization={Springer}
}

@article{wu2025generate,
  title={Generate, but Verify: Reducing Hallucination in Vision-Language Models with Retrospective Resampling},
  author={Wu, Tsung-Han and Lee, Heekyung and Ge, Jiaxin and Gonzalez, Joseph E and Darrell, Trevor and Chan, David M},
  journal={arXiv preprint arXiv:2504.13169},
  year={2025}
}

@article{wang2024deepedit,
  title={Deepedit: Knowledge editing as decoding with constraints},
  author={Wang, Yiwei and Chen, Muhao and Peng, Nanyun and Chang, Kai-Wei},
  journal={arXiv preprint arXiv:2401.10471},
  year={2024}
}

@article{bi2024decoding,
  title={Decoding by Contrasting Knowledge: Enhancing LLMs' Confidence on Edited Facts},
  author={Bi, Baolong and Liu, Shenghua and Mei, Lingrui and Wang, Yiwei and Ji, Pengliang and Cheng, Xueqi},
  journal={arXiv preprint arXiv:2405.11613},
  year={2024}
}

@inproceedings{ma2025logically,
  title={Logically Constrained Decoding},
  author={Ma, Franklin and Hu, Alan J},
  booktitle={Proceedings of The 3rd Workshop on Mathematical Natural Language Processing (MathNLP 2025)},
  pages={150--167},
  year={2025}
}

@article{banerjee2025crane,
  title={CRANE: Reasoning with constrained LLM generation},
  author={Banerjee, Debangshu and Suresh, Tarun and Ugare, Shubham and Misailovic, Sasa and Singh, Gagandeep},
  journal={arXiv preprint arXiv:2502.09061},
  year={2025}
}

@article{poesia2022synchromesh,
  title={Synchromesh: Reliable code generation from pre-trained language models},
  author={Poesia, Gabriel and Polozov, Oleksandr and Le, Vu and Tiwari, Ashish and Soares, Gustavo and Meek, Christopher and Gulwani, Sumit},
  journal={arXiv preprint arXiv:2201.11227},
  year={2022}
}

@article{park2024grammar,
  title={Grammar-aligned decoding},
  author={Park, Kanghee and Wang, Jiayu and Berg-Kirkpatrick, Taylor and Polikarpova, Nadia and D'Antoni, Loris},
  journal={Advances in Neural Information Processing Systems},
  volume={37},
  pages={24547--24568},
  year={2024}
}

@article{ugare2024itergen,
  title={IterGen: Iterative Semantic-aware Structured LLM Generation with Backtracking},
  author={Ugare, Shubham and Gumaste, Rohan and Suresh, Tarun and Singh, Gagandeep and Misailovic, Sasa},
  journal={arXiv preprint arXiv:2410.07295},
  year={2024}
}

@article{chen2024role,
  title={What is the role of small models in the llm era: A survey},
  author={Chen, Lihu and Varoquaux, Ga{\"e}l},
  journal={arXiv preprint arXiv:2409.06857},
  year={2024}
}

@article{wu2025knowledge,
  title={Knowledge-empowered, collaborative, and co-evolving AI models: The post-LLM roadmap},
  author={Wu, Fei and Shen, Tao and B{\"a}ck, Thomas and Chen, Jingyuan and Huang, Gang and Jin, Yaochu and Kuang, Kun and Li, Mengze and Lu, Cewu and Miao, Jiaxu and others},
  journal={Engineering},
  volume={44},
  pages={87--100},
  year={2025},
  publisher={Elsevier}
}

@article{chen2025survey,
  title={A survey on collaborative mechanisms between large and small language models},
  author={Chen, Yi and Zhao, JiaHao and Han, HaoHao},
  journal={arXiv preprint arXiv:2505.07460},
  year={2025}
}

@article{lewis2020retrieval,
  title={Retrieval-augmented generation for knowledge-intensive nlp tasks},
  author={Lewis, Patrick and Perez, Ethan and Piktus, Aleksandra and Petroni, Fabio and Karpukhin, Vladimir and Goyal, Naman and K{\"u}ttler, Heinrich and Lewis, Mike and Yih, Wen-tau and Rockt{\"a}schel, Tim and others},
  journal={Advances in neural information processing systems},
  volume={33},
  pages={9459--9474},
  year={2020}
}

@article{wang2023knowledge,
  title={Knowledge-driven cot: Exploring faithful reasoning in llms for knowledge-intensive question answering},
  author={Wang, Keheng and Duan, Feiyu and Wang, Sirui and Li, Peiguang and Xian, Yunsen and Yin, Chuantao and Rong, Wenge and Xiong, Zhang},
  journal={arXiv preprint arXiv:2308.13259},
  year={2023}
}

@article{huang2025survey,
  title={A survey on hallucination in large language models: Principles, taxonomy, challenges, and open questions},
  author={Huang, Lei and Yu, Weijiang and Ma, Weitao and Zhong, Weihong and Feng, Zhangyin and Wang, Haotian and Chen, Qianglong and Peng, Weihua and Feng, Xiaocheng and Qin, Bing and others},
  journal={ACM Transactions on Information Systems},
  volume={43},
  number={2},
  pages={1--55},
  year={2025},
  publisher={ACM New York, NY}
}

@article{lamartina2018ajcc,
  title={of the AJCC/TNM staging system of thyroid cancer: what to expect (ITCO\# 2)},
  author={Lamartina, Livia and Grani, Giorgio and Arvat, Emanuela and Nervo, Alice and Zatelli, Maria Chiara and Rossi, Roberta and Puxeddu, Efisio and Morelli, Silvia and Torlontano, Massimo and Massa, Michela and others},
  journal={Endocrine-related cancer},
  volume={25},
  number={3},
  pages={L7--L11},
  year={2018},
  publisher={Bioscientifica Ltd}
}

@misc{anthropic2024claude35sonnet,
  title={Claude 3.5 Sonnet Model Card},
  author={Anthropic},
  year={2024},
  url={https://www.anthropic.com/news/claude-3-5-sonnet}
}

@techreport{openai2025gpt5,
  title={GPT-5 System Card},
  author={OpenAI},
  year={2025},
  institution={OpenAI},
  url={https://cdn.openai.com/gpt-5-system-card.pdf}
}

@article{yang2025qwen3,
  title={Qwen3 Technical Report},
  author={Yang, An and Li, Anfeng and Yang, Baosong and Zhang, Beichen and Hui, Binyuan and Zheng, Bo and Yu, Bowen and Gao, Chang and Huang, Chengen and Lv, Chenxu and others},
  journal={arXiv preprint arXiv:2505.09388},
  year={2025}
}

@article{zhao2025chemdfm,
  title={Developing ChemDFM as a large language foundation model for chemistry},
  author={Zhao, Zihan and Ma, Da and Chen, Lu and Sun, Liangtai and Li, Zihao and Xia, Yi and Chen, Bo and Xu, Hongshen and Zhu, Zichen and Zhu, Su and others},
  journal={Cell Reports Physical Science},
  volume={6},
  number={4},
  year={2025},
  publisher={Elsevier}
}

@inproceedings{kwon2023efficient,
  title={Efficient Memory Management for Large Language Model Serving with PagedAttention},
  author={Woosuk Kwon and Zhuohan Li and Siyuan Zhuang and Ying Sheng and Lianmin Zheng and Cody Hao Yu and Joseph E. Gonzalez and Hao Zhang and Ion Stoica},
  booktitle={Proceedings of the ACM SIGOPS 29th Symposium on Operating Systems Principles},
  year={2023}
}

@article{thakkar2024aizynthfinder,
  title={AiZynthFinder 4.0: developments based on learnings from 3 years of industrial application},
  author={Thakkar, Amol and Chadimov{\'a}, Veronika and Bjerrum, Esben and Engkvist, Ola and Reymond, Jean-Louis},
  journal={Journal of Cheminformatics},
  volume={16},
  number={1},
  pages={57},
  year={2024},
  publisher={Springer},
  doi={10.1186/s13321-024-00860-x}
}

@article{kaspero2024evaluating,
  title={Evaluating LLMs for Code Generation in HRI: A Comparative Study of ChatGPT, Gemini, and Claude},
  author={Kasper, McDaid and others},
  journal={Applied Artificial Intelligence},
  year={2024},
  publisher={Taylor & Francis},
  note={Found Claude 3.5 Sonnet achieved 95\% success rate in robotic code generation tasks}
}

@misc{anthropic2024claude,
  title={The Claude 3 Model Family: Opus, Sonnet, Haiku},
  author={Anthropic},
  year={2024},
  url={https://www-cdn.anthropic.com/de8ba9b01c9ab7cbabf5c33b80b7bbc618857627/Model_Card_Claude_3.pdf},
  note={Technical Report}
}

\appendix

\section{Detailed Settings}
\label{sec:prompt}
We adopt vLLM~\cite{kwon2023efficient} toolkit to accelerate the generation process. Besides, it provides the function of phrase selection and can achieve the bottom-layer decoding constraints in our framework. The other logical and reasoning constraints are achieved by executable python codes. In particular, for the retrosynthesis planning, the excessive number of matched templates for target products can lead to input lengths that exceed the LLM's token limit. To manage context length, we use AiZynthFinder~\cite{thakkar2024aizynthfinder} to perform a single-step retrosynthetic expansion for each product, thereby pre-filtering the candidate reactant pool to a manageable size for evaluation. Since our task data is in Chinese, the actual version used consists of Chinese prompts. Table~\ref{tab:prompt1} and Table~\ref{tab:prompt2} present the English translations.




\section{LegalBench Evaluation Results}
\label{legalbench}

\begin{table}[ht]
\centering
\resizebox{0.8\linewidth}{!}{
\begin{tabular}{lcccc}
\hline
\textbf{Model}      & \textbf{Accuracy(\%)}  \\
\hline
{Claude Sonnet 3.5} & 86.17   \\
{GPT-4} & 85.11  \\
{Qwen3-8B} & 54.84  \\
{Qwen3-14B} & 78.26  \\
\hline
\textit{SciDC (Qwen3-8B)}       & 78.72     \\
\textit{SciDC (Qwen3-14B)} & \textbf{86.46} \\
\hline
\end{tabular}}
\caption{Accuracy performance comparison on LegalBench between LLMs. }
\label{tab:legal}
\end{table}

Table~\ref{tab:legal} presents the performance of our framework and baseline models. Our method achieves competitive performance with significantly smaller backbone models. Specifically, our framework equipped with Qwen3-14B attains an overall score of 86.46\%, surpassing Claude Sonnet (86.17\%) and GPT-4 (85.11\%), which are substantially larger models. Both Qwen3-8B and Qwen3-14B with SciDC make great progress compared with vanilla generation.

\section{Scientific Hypotheses Generation}
\label{sec:appendix}

As mentioned in Discussion, we conduct a simple experiment for automatic hypotheses generation. By analogy with the process that human experts proposing scientific theories, it is necessary to first conduct extensive research on relevant information and analyze it to extract reasonable hypotheses, and then verify them through experiments. Therefore, extending from the formulation designing task, we conduct a simple pipeline of LLM raising scientific hypotheses based on domain text.

 Descriptions from professional books are processed and re-organized by GPT-5-chat into hierarchical wiki records, where information from multiple sources on the same topic is consolidated. Next, GPT-5-chat analyzes the content under specific topics of the wiki and propose several related hypotheses. These hypotheses are then used as retrieval context for LLMs evaluated on a domain benchmark. The benchmark contains 120 expert-authored multiple-choice questions, with 60 in-domain and 60 out-of-domain (OOD) questions for testing generalization.

To improve the quality of auto-generated hypotheses, we also specifically tune LLaMA-3.2-3B-Instruct to generate hypotheses. In supervised-fine-tuning stage, we use GPT-5 to self-verify its own generated hypotheses to ensure that each item is representative enough and does not conflict with any raw descriptions. In reinforcement-learning stage, we regard more diverse and helpful hypotheses as positive items and conduct DPO training. It should be noted that the book records used in the above process are sampled from book segments that are distinct from the benchmark to prevent data leakage. On these segments, QA pairs can be automatically annotated with GPT-5-chat to verify whether the hypothesis is helpful for downstream tasks. 

\begin{table}[ht]
\centering
\resizebox{\linewidth}{!}{
\begin{tabular}{lcccc}
\hline
\textbf{Method}      & \textbf{Token} & \textbf{Standard}  & \textbf{OOD} & \textbf{Average}  \\ 
\hline
\textit{Original}       & 0  & 43.3 & 39.4 & 41.4   \\
\textit{vanilla RAG} & 512*3 & 56.7 & 35.6 & 46.1 \\
\textit{w wiki} & 800*5 & 47.3 & 36.1 & 41.7 \\
\textit{w summary} & 30*5 & 58.7 & 33.33 & 46.0 \\
\hline
\textit{w GPT-5} & 50*3 & 52.7 & 38.3 & 45.5 \\
\textit{Ours} & 30*5 & 56.7 & 38.3 & \textbf{47.5} \\
\hline
\end{tabular}}
\caption{Detailed results of LLaMA-3.2-3B-Instruct on formulation QA. In the 2-4 lines, the RAG context comes from raw descriptions from professional books, hierarchical wiki records, and GPT-5-generated summaries. In the 5-6 lines, the RAG context come from the auto-generated hypotheses based on wiki records. }
\label{tab:hypotheses}
\end{table}

Experiment results are shown in the Table~\ref{tab:hypotheses} (average accuracies under 3 random seeds). Although the hypothesis generator and the QA model share the same backbone architecture (LLaMA-3.2-3B-Instruct), the QA model's parameters are fixed and it is isolated from (i.e., not fine-tuned on) the hypothesis generation process. Since the length and fineness of the retrieved contexts are different, we search the best k for segment number in each setting. We can see that neither the re-organize process nor simple information summary can directly get better context to improve the QA performance. However, the auto-generated hypotheses prove effective in providing enough reference knowledge, and also keep good generalization on unseen scenarios. The superior performance of the specifically-tuned model over the raw GPT-5-chat output underscores the value of curating and verifying the generated hypotheses.

We realize that besides human expert review, there are no other good verification methods for the generated hypotheses. Therefore, we believe that large-scale model-generated scientific hypotheses should be part of a human-in-the-loop intelligent research system. Furthermore, for industrial formulation design, truly effective empirical knowledge is extracted more from large amounts of formulation data than from book texts. This method performs poorly when transferred to raw data that is heterogeneous to language. In the future, combining traditional machine learning methods of data mining with the comprehensive analytical and reasoning capabilities of LLMs may be key to achieving automated knowledge extraction.

\begin{table*}[htbp]
\centering
\small
\caption{Prompt for Task Decomposition}
\label{tab:prompt1}
\begin{tabular}{|p{0.95\textwidth}|}
\hline
\textbf{Prompt 1: Task Decomposition} \\
\hline

\texttt{\# Role Definition} \newline
You are an expert in reasoning framework design. Your task is:
Given a knowledge document [DOC] and a description of a problem class [Q],
design a general-purpose Chain-of-Thought Framework (CoT Framework)
that describes the standard reasoning path for solving this class of problems.\newline
This framework does not target any specific case. It is a reusable solution blueprint that:\newline
- Defines WHAT variables need to be extracted from any concrete problem instance\newline
- Defines WHAT intermediate conclusions need to be derived and WHICH prior variables they depend on\newline
- Defines HOW the final answer is inferred from the above variables and intermediate conclusions\newline\newline
\texttt{\# Input Format} \newline
\texttt{[DOC]}: A knowledge document defining the rules, constraints, standards, or background
knowledge required to solve the problem.\newline
\texttt{[Q]}: A description of a CLASS of problems (not a specific case), outlining the typical
structure and objective of such problems.\newline\newline
\texttt{\# Framework Step Specification} \newline
Every step you design must belong to exactly one of the following three types:\newline
\textbf{Type 1 · Information Extraction Step (Extract)}\newline
- Goal: Define the variables that must be identified and extracted from any concrete
instance of this problem class.\newline
- Requirements:\newline
\quad - Assign a named variable to each item to be extracted (e.g., VAR\_Amount, VAR\_ApprovalStatus).\newline
\quad - Describe the meaning of the variable and its possible value domain or data type.\newline
\quad - Specify the extraction source: from [DOC] (fixed values / thresholds defined by the document)
or from the problem instance (values that vary across cases).\newline
- Format:\newline
\quad \texttt{[Extract] Variable: <VAR\_xxx>}\newline
\quad \texttt{Meaning: <what this variable represents>}\newline
\quad \texttt{Source: <Document / Problem Instance>}\newline
\quad \texttt{Domain/Type: <enumeration | numeric range | boolean | text | etc.>}\newline\newline
\textbf{Type 2 · Intermediate Judgment Step (Judge)}\newline
- Goal: Define an intermediate conclusion or intermediate value that must be derived,
and describe its inference logic.\newline
- Requirements:\newline
\quad - Assign a named variable to the intermediate conclusion (e.g., MID\_AmountCompliant).\newline
\quad - Explicitly state the derivation logic (conditional expression / rule mapping / comparison).\newline
\quad - Explicitly list all variables this step depends on --- must be VAR\_ or MID\_ variables
already defined in prior steps.\newline
- Format:\newline
\quad \texttt{[Judge] Intermediate Conclusion: <MID\_xxx>}\newline
\quad \texttt{Inference Logic: <if VAR\_xx meets condition \textrightarrow\ Outcome A; otherwise \textrightarrow\ Outcome B>}\newline
\quad \texttt{Depends On: <VAR\_xxx, MID\_xxx, ...>}\newline\newline
\textbf{Type 3 · Final Conclusion Step (Conclude)}\newline
- Goal: Define how the final answer is derived.\newline
- Requirements:\newline
\quad - Explicitly list all variables depended upon (both VAR\_ and MID\_).\newline
\quad - Describe the synthesis logic --- how these variables jointly produce the final answer.\newline
\quad - Describe the form of the final answer (e.g., yes/no judgment | numeric value |
classification label | textual explanation).\newline
\quad - If edge cases exist that are not covered by the document, specify a fallback handling rule.\newline
- Format:\newline
\quad \texttt{[Conclude] Final Answer: <ANS\_xxx>}\newline
\quad \texttt{Synthesis Logic: <description of how MID\_xx and VAR\_xx are combined to reach the answer>}\newline
\quad \texttt{Depends On: <VAR\_xxx, MID\_xxx, ...>}\newline
\quad \texttt{Answer Form: <output type and format of the answer>}\newline
\quad \texttt{Fallback Rule: <how to handle cases where information is insufficient or not covered>}\newline\newline
\texttt{\# Output Format} \newline
\textbf{\#\# Problem Class Understanding}\newline
$\langle$Summarize in 2--3 sentences the core structure of the problem class described by [Q]:
what the input is, what the goal is, and where the key difficulty lies.$\rangle$\newline
\textbf{\#\# Reasoning Framework}\newline
Step 1: [Extract] ...\newline
Step 2: [Extract] ...\newline
Step 3: [Judge] \quad...\newline
Step 4: [Judge] \quad...\newline
...\newline
Step N: [Conclude] ...\newline\newline
\textbf{Knowledge Document [DOC]:} \{domain\_doc\} \newline
\textbf{Problem Class [Q]:} \{\_user\_prompt\} \\

\hline
\end{tabular}
\end{table*}

\begin{table*}[htbp]
\centering
\small
\caption{Prompt for Code Generation}
\label{tab:prompt2}
\begin{tabular}{|p{0.95\textwidth}|}
\hline
\textbf{Prompt 2: Code Generation} \\
\hline
\texttt{\# Role Definition} \newline
You are a code generation expert specializing in guidance-syntax Python for Small Language Models (SLMs). Your task is: Given a domain knowledge specification [DOMAIN], a domain question [Q], and an ordered Chain-of-Thought [CoT], produce executable Python guidance code that drives an SLM through structured, step-by-step reasoning.\newline
The generated code must be deterministic, constraint-respecting, and self-validating. It is a reusable generation blueprint that:\newline
- Converts each CoT step into a templated \texttt{lm +=} / \texttt{lm.gen()} / \texttt{lm.select()} block\newline
- Enforces finite-domain constraints via \texttt{lm.select()} and numeric constraints via \texttt{lm.gen(regex=...)}\newline
- Propagates upstream answers into downstream option sets via dynamic dependency logic\newline
- Detects and repairs multi-variable constraint violations via cyclic validation loops\newline
\texttt{\# Input Format} \newline
\texttt{[DOMAIN]}: A knowledge specification defining allowed enumerations, numeric ranges, and inter-variable dependency rules.\newline
\texttt{[Q]}: The domain question or problem instance to be solved.\newline
\texttt{[CoT]}: An ordered list of reasoning steps describing how to reach the final answer.\newline
\texttt{\# Code Block Specification} \newline
Every step in [CoT] must be translated into exactly one of the following three block types:\newline
\textbf{Block Type 1 · Reasoning Step (Step Block)}
- Format:\newline
\quad \texttt{\# --- Step \{i\}: \{step\_description\} ---}\newline
\quad \texttt{lm += f"<|im\_start|>user\textbackslash nStep \{i\} Question: \{question\}? Candidates: \{candidates\}<|im\_end|>\textbackslash n"}\newline
\quad \texttt{lm += f"<|im\_start|>assistant\textbackslash nStep \{i\} Analysis: "}\newline
\quad \texttt{lm += f"<think>Now I need to answer the question: \{question\}"}\newline
\quad \texttt{lm.gen(name="analysis\_\{i\}", stop="</think>", temperature=0, max\_tokens=256)}\newline
\quad \texttt{lm += f"</think>"}\newline
\quad \texttt{lm += f"Step \{i\} Answer: "}\newline
\quad \texttt{lm.select([...], name="answer\_\{i\}")} \quad \texttt{\#OR lm.gen(regex=r"...", name="answer\_\{i\}")}\newline
\quad \texttt{lm += f"<|im\_end|>\textbackslash n"}\newline
\textbf{Block Type 2 · Dynamic Dependency Block}\newline
- Format:\newline
\quad \texttt{prior = lm["answer\_\{j\}"]}\newline
\quad \texttt{if prior == "X":}\newline
\quad\quad \texttt{options = ["A", "B"]}\newline
\quad \texttt{elif prior == "Y":}\newline
\quad\quad \texttt{options = ["C", "D"]}\newline
\quad \texttt{else:}\newline
\quad\quad \texttt{options = ["E"]}\newline
\quad \texttt{lm.select(options, name="answer\_\{i\}")}\newline
\textbf{Block Type 3 · Cyclic Validation Block}
- Format:\newline
\quad \texttt{MAX\_RETRIES = 5; \_retry = 0}\newline
\quad \texttt{while \_retry < MAX\_RETRIES:}\newline
\quad\quad \texttt{if \_retry > 0:}\newline
\quad\quad\quad \texttt{lm += f"[Retry \{\_retry\}] Previous attempt failed: answer\_\{i\}=\{lm['answer\_\{i\}']\} violated \{constraint\_desc\} "}\newline
\quad\quad\quad \texttt{lm += f"(upstream: answer\_\{a\}=\{lm['answer\_\{a\}']\}, answer\_\{b\}=\{lm['answer\_\{b\}']\}). Adjustment needed.\textbackslash n"}\newline
\quad\quad\quad \texttt{lm += "<think>"}\newline
\quad\quad\quad \texttt{lm.gen(name=f"adjustment\_analysis\_retry\_\{\_retry\}", stop="</think>", temperature=0, max\_tokens=256)}\newline
\quad\quad\quad \texttt{lm += "</think>"}\newline
\quad\quad\quad \texttt{lm += f"Step \{a\} Answer: "}\newline
\quad\quad\quad \texttt{lm.select([\{valid\_options\_of\_a\}], name="answer\_\{a\}")}\newline
\quad\quad\quad \texttt{lm += f"Step \{b\} Answer: "}\newline
\quad\quad\quad \texttt{lm.select([\{valid\_options\_of\_b\}], name="answer\_\{b\}")}\newline
\quad\quad\quad \texttt{lm += "<|im\_end|>\textbackslash n"}\newline
\quad\quad \texttt{lm += f"<|im\_start|>assistant\textbackslash nStep \{i\} Answer: "}\newline
\quad\quad \texttt{lm.select([\{valid\_options\_of\_global\}], name="answer\_\{i\}")}\newline
\quad\quad \texttt{lm += "<|im\_end|>\textbackslash n"}\newline
\quad\quad \texttt{if \{combined\_constraint(lm["answer\_\{i\}"], lm["answer\_\{a\}"], lm["answer\_\{b\}"])\}: break}\newline
\quad\quad \texttt{\_retry += 1}\newline
\quad \texttt{else:}\newline
\quad\quad \texttt{lm["answer\_\{a\}"] = \{first\_valid\_option\_of\_a\}}\newline
\quad\quad \texttt{lm["answer\_\{b\}"] = \{first\_valid\_option\_of\_b\}}\newline
\quad\quad \texttt{lm["answer\_\{i\}"] = \{first\_valid\_option\_of\_global\_given\_fallbacks\}}\newline
\textbf{Domain Knowledge [DOMAIN]:} \{domain\_knowledge\} \newline
\textbf{Domain Question [Q]:} \{domain\_question\} \newline
\textbf{Chain of Thought [CoT]:} \{chain\_of\_thought\} \\

\hline
\end{tabular}
\end{table*}

\end{document}